\documentclass[10pt, a4paper]{article}

\usepackage[final]{lrec2026} 
\usepackage{float}
\usepackage{graphicx}
\usepackage{tabularx,booktabs,array,xcolor,amsmath,multirow,subcaption}
\title{Reason2Decide: Rationale-Driven Multi-Task Learning}
\name{\textbf{H M Quamran Hasan$^1$, Housam Khalifa Bashier$^1$, Jiayi Dai$^1$,} \\ 
      \large \textbf{Mi-Young Kim$^2$, Randy Goebel$^1$}}

\address{$^1$University of Alberta, Edmonton, Alberta, Canada \\
         $^2$University of Alberta, Camrose, Alberta, Canada \\
         {\{hmquamra, khalifab, dai1, miyoung2, rgoebel\}@ualberta.ca}}

\abstract{
Despite the wide adoption of Large Language Models (LLM)s, clinical decision support systems face a critical challenge: achieving high predictive accuracy while generating explanations aligned with those predictions. Current approaches suffer from exposure bias, leading to misaligned explanations. We propose Reason2Decide, a two-stage training framework that addresses key challenges in self-rationalization, including exposure bias and task separation. In Stage-1, our model is trained on rationale generation, while in Stage-2, we jointly train on label prediction and rationale generation, applying scheduled sampling to gradually transition from conditioning on gold labels to model predictions.
We evaluate Reason2Decide on three medical datasets, including a proprietary triage dataset and public biomedical QA datasets. Across model sizes, Reason2Decide outperforms other fine-tuned baselines and some zero-shot LLMs in prediction (F1) and rationale fidelity (BERTScore, BLEU, LLM-as-a-Judge).
In triage, Reason2Decide is rationale source-robust across LLM-generated, nurse-authored, and nurse-post-processed rationales. In our experiments, while using only LLM-generated rationales in Stage-1, Reason2Decide outperforms other fine-tuned variants. This indicates that LLM-generated rationales are suitable for pretraining models, reducing reliance on human annotations. Remarkably, Reason2Decide achieves these gains with models $40\times$ smaller than contemporary foundation models, making clinical reasoning more accessible for resource-constrained deployments while still providing explainable decision support.
 \\ \newline  }

\begin{document}
\maketitleabstract
\section{Introduction}

The integration of reasoning capabilities with prediction tasks has been a critical research problem in natural language processing (NLP). Existing state-of-the-art large language models (LLMs) struggle to balance high predictive accuracy while also generating human-interpretable explanations \cite{niu-etal-2025-knowledge}. Although LLMs have demonstrated strong performance on various question-answering benchmarks \cite{hendrycks2021measuringmassivemultitasklanguage, srivastava2023imitationgamequantifyingextrapolating}, their ability to provide rationales that align with their predictions remains limited. This limitation is substantially amplified in healthcare where explanations are essential for trust and adoption.

Current approaches for rationale generation face two fundamental challenges. First, the exposure bias problem \cite{schmidt-2019-generalization}, which arises when models are trained to generate rationales conditioned only on ground-truth labels, but need to explain their own potentially incorrect predictions during inference. Second, the task separation problem, which occurs when prediction and rationale generation are treated independently rather than as mutually supporting tasks \cite{narang2020wt5trainingtexttotextmodels}. While multi-task learning has shown promising performance, existing methods have failed to address the train-test discrepancy: models exclusively learn to explain gold-standard labels during training, leaving them unprepared for rationalizing their own predictions during inference.  

Recent work has acknowledged these challenges and explored various approaches to improve model interpretability. For example, ``Chain-of-Thought'' (CoT) prompting  \cite{NEURIPS2022_9d560961} demonstrates that step-by-step reasoning can improve complex reasoning; similarly, self-consistency techniques \cite{wang2023selfconsistencyimproveschainthought} show the value of aggregating multiple reasoning paths. However, these methods primarily focus on inference time strategies rather than addressing the fundamental training dynamics that help establish more robust reasoning capabilities. Knowledge distillation approaches \cite{hsieh-etal-2023-distilling} and rationale-augmented training \cite{lampinen2022tellwhyexplanationssupport} have also shown benefits, but do not train models to generate explanations conditioned on their own predictions during training.  

\begin{figure*}[h!]
\begin{center}
\includegraphics[width=0.8\textwidth]{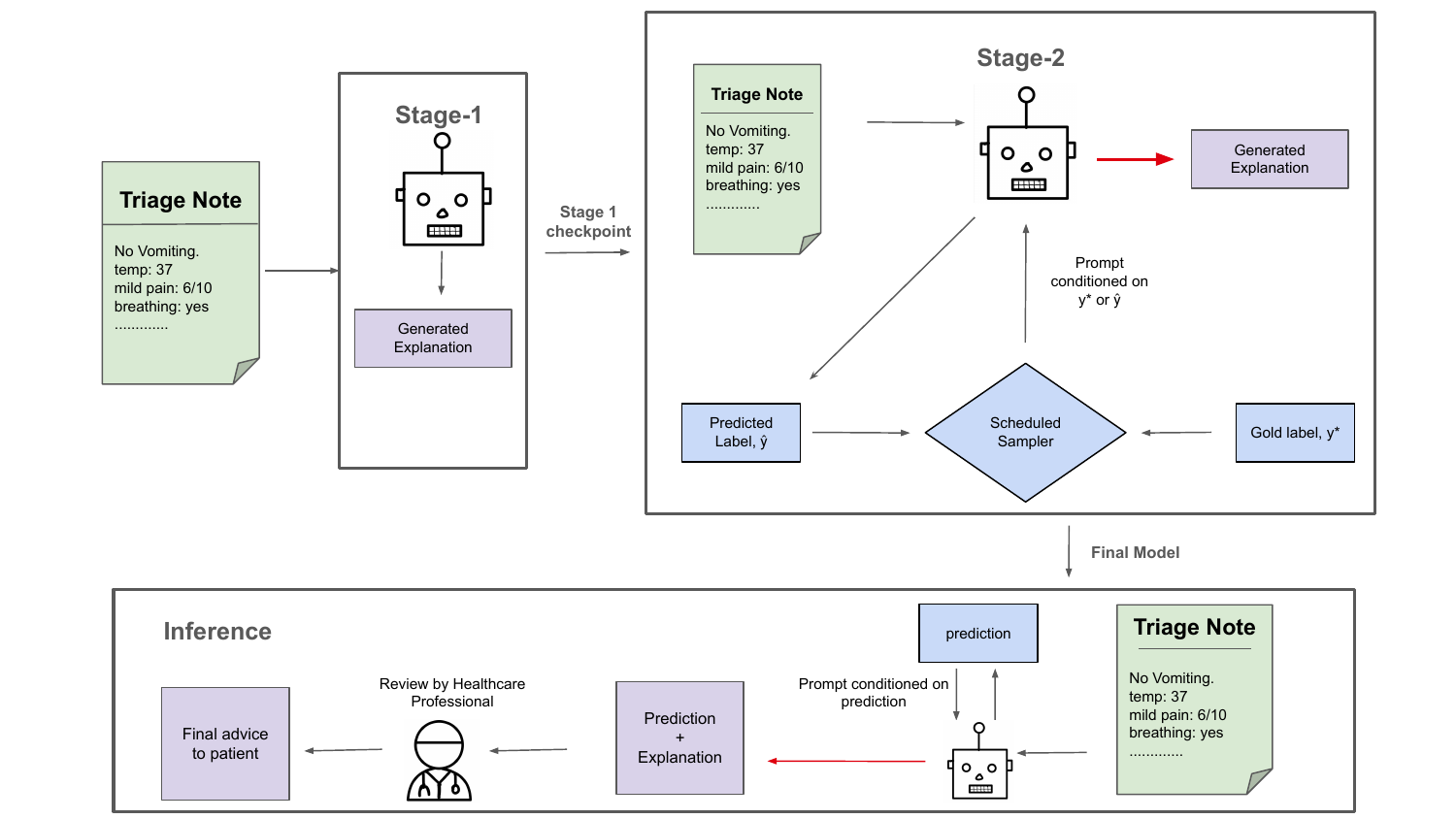}

\caption{Overview of Reason2Decide. Stage-1 trains rationale generation. Stage-2 jointly predicts labels and generates label-conditioned explanations with task-level scheduled sampling. A single T5 model is used throughout; inference conditions explanations on the model’s own prediction.}
\label{fig.methodology}
\end{center}
\end{figure*}

Here we propose Reason2Decide (Figure \ref{fig.methodology}), a framework which trains a single model to jointly predict and generate rationales. Our approach consists of a two-stage training regime, based on insights from curriculum learning \cite{10.1145/1553374.1553380} and scheduled sampling \cite{10.5555/2969239.2969370}. The idea is to first learn to model explanation fundamentals and then jointly produce predictions and rationales, where we gradually shift from gold-label conditioning to self-conditioning. Through this process, the model learns to explain its own predictions. This addresses the exposure bias problem while promoting alignment between predictions and rationales. 
The main contributions of this paper are:

\begin{itemize}
\item We propose Reason2Decide, a two-stage training framework for LLMs that first learns the fundamentals of explanation generation from rationales and then jointly optimizes predictions and rationales in a multi-task setup.
\item We introduce a scheduled sampling mechanism that gradually transitions from gold labels to predicted label conditioning, mitigating exposure bias in self-rationalization. 

\end{itemize}

\section{Related Work}
Our work builds upon research in rationale generation, multi-task learning, and existing methods to mitigate exposure bias. 

\subsection{Rationale Generation and Explainable AI}
A primary goal of Explainable AI (XAI) is making model predictions transparent and interpretable. The development of early XAI methods focused on post-hoc explanation generation by analyzing learned model internal attributes or feature importance \cite{ribeiro-etal-2016-trust,lundberg2017unifiedapproachinterpretingmodel}. Recently, the field has shifted towards \emph{self-explaining models} that can learn to generate rationales simultaneously with their predictions \cite{camburu2018esnlinaturallanguageinference}. Our work aligns with the latter. For instance, \citet{narang2020wt5trainingtexttotextmodels} fine-tuned T5 \cite{raffel2023exploringlimitstransferlearning} models to generate explanations by treating them as a separate task. Similarly, \citet{lampinen2022tellwhyexplanationssupport} demonstrate that incorporating natural language explanations during training can improve reasoning and generalization. However, these approaches often treat the explanation as a secondary output, which creates a separation between the prediction and its explanation. In contrast, our Reason2Decide framework is designed to align the predictive task with the explanatory one, to help ensure that the explanation is an integral part of the reasoning process.

\subsection{Multi-Task Learning and Knowledge Distillation}
Multi-Task Learning aims to improve generalization by leveraging shared representations across related tasks \cite{ruder2017overviewmultitasklearningdeep,10.1023/A:1007379606734}. In NLP, models like T5 are considered multi-task learners because they combine diverse problems into a unified text-to-text format. We similarly use this foundation but specifically focus on the alignment between prediction and rationales. Closely related work includes knowledge distillation approaches that use rationales. For example, \citet{hsieh-etal-2023-distilling}  distills the capabilities of a larger teacher model that generates step-by-step rationales into a smaller student model. While this approach can be effective in improving predictive accuracy, it does not explicitly enforce prediction–explanation coordination. The student model benefits from teacher rationales but is not trained to ensure that its own predictions and explanations are self-consistent. Our method addresses this weakness by explicitly training the coordination between prediction and rationale generation.

\subsection{Mitigating Exposure Bias and Scheduled Sampling}
Exposure bias in self-rationalization occurs when models are trained to generate explanations conditioned on gold labels but at inference time must justify their own predictions. Scheduled sampling \cite{10.5555/2969239.2969370} addresses the issue by gradual replacement of teacher-forced tokens with model-generated ones. Here, we adapt this principle to the task level, thus treating the model's predicted label as the conditioning context for rationale generation. Our mechanism systematically transitions from using gold labels to the model's own predictions, directly mitigating this bias and preparing models for real-world deployment.

\section{Methodology}

Our method addresses the problem of combined prediction and rationale generation, where given an input $x$ (clinical note or biomedical question), a model must predict a discrete label $y \in \mathcal{Y}$ and generate a free-text rationale $r$ which justifies the prediction.

We employ a single encoder-decoder architecture $f_\theta$ (T5 variants) without modifications to its core components including the number of attention heads, hidden layer dimensions, relative positional encodings, and activation functions which are identical to the original specification. T5's text-to-text formulation is ideal for our rationale-driven paradigm, as it naturally supports autoregressive decoding for both discrete label prediction (e.g., ``Yes", ``No", or ``Go to ED Now.") and free-form rationale generation from task-specific prompts. This allows seamless multi-tasking with explanations under a single objective, unlike encoder-only models like BERT \cite{devlin2019bertpretrainingdeepbidirectional}, which require separate classification heads and lack native support for conditional explanation generation. 

Our training configuration is designed to handle both tasks: predict a discrete label and generate a free-text rationale. The model first predicts the label as \texttt{predict:}$x$ $\mapsto$ $\hat{y}$, then employs the label to condition the rationale as: \texttt{given label:}$y^{*}$ \texttt{or} $\hat{y}$\texttt{, explain:}$x$ $\mapsto$ $\hat{r}$, where $y^{*}$ is the gold label, $\hat{y}$ is the model's predicted label, and $\hat{r}$ is the generated rationale.
 


\subsection{Stage-1: Rationale Foundation Training}
Previous work has shown that domain-specific pretraining significantly improves model performance in specialized tasks \cite{lewis-etal-2020-pretrained,10.1093/bioinformatics/btz682}. With this motivating background, our first stage teaches explanation fundamentals by training the model to generate rationales. Let $\theta$ denote the parameters of the T5 model. Given a gold rationale $r^*$, we optimize:
\[
\mathcal{L}_{\text{stage1}} = -\log P_\theta(r^* \mid \texttt{explain:} x)
\]
where \( P_\theta(r^* \mid \cdot) \) denotes the autoregressive probability of the full rationale token sequence. The best model checkpoint is used to initialize Stage-2.


This two-stage structure follows a curriculum-inspired design, where the model first learns explanation fundamentals before being required to justify its own predictions.

\subsection{Stage-2: Joint Optimization of Prediction and Explanation with Task-Level Scheduled Sampling}

The second stage jointly optimizes prediction and explanation, starting with gold labels and gradually shifting to predicted labels for conditioning.

\paragraph{Prediction Task: } The model treats label prediction as a text generation task, optimized via cross-entropy loss:
\[
\mathcal{L}_{\text{pred}} = -\log P_\theta\big(y^* \,\big|\, \texttt{predict:}x\big)
\]
where \( P_\theta(y^* \mid \cdot) \) denotes the autoregressive probability of the full gold label sequence. 

\paragraph{Explanation with Task-Level Scheduled Sampling: }To mitigate exposure bias in self-rationalization, we introduce a scheduled sampling mechanism.
Unlike prior token-level scheduled sampling \cite{10.5555/2969239.2969370}, which replaces target tokens within sequences, we apply it at the task level by switching the conditioning context from gold labels to model-predicted labels.
For instance, instead of gradually replacing reference tokens in a generated sequence, our method gradually replaces reference labels used for conditioning the rationale generation.
 
 For each example we condition on either the gold label $y^{*}$ or the model prediction $\hat{y}$ (greedily decoded during training):
\[
\tilde{y} = 
\begin{cases} 
y^* & \text{with probability } 1 - \pi_{t} \\
\hat{y} & \text{with probability } \pi_{t}
\end{cases}
\]

where $\pi_{t}$ follows a linear schedule. With total training steps $T$, warm-up phase $w = 0.05T$, transition phase $m = 0.60T$, for step $t$:

\[
\pi_{t} =
\begin{cases}
0, & 0 \le t < w, \\[4pt]
\min\left(0.9, \dfrac{t - w}{m}\right), & w \le t < w + m, \\[10pt]
0.9, & t \ge w + m.
\end{cases}
\]

The ceiling of $0.9$ was a design choice to prevent the model from fully relying on self-generated labels during training, in order to help avoid error amplification from incorrect predictions. The hyperparameters ($0.05$ and $0.6$) were determined through hyperparameter optimization using a few sets of predefined tuples, and are motivated by two key considerations:

\paragraph{Warm-up Phase:} During $0<=t<w$, we linearly increase $\alpha_{t}$ from $0$ to $0.7$. This is to prioritize explanation loss $\mathcal{L}_{\text{expl}}$ over prediction loss $\mathcal{L}_{\text{pred}}$, to provide a smoother transition from single-task to multi-task optimization. After this phase $\alpha_{t}$ is held constant at $0.7$ for the rest of the training. We heuristically selected $0.7$ to balance both tasks, with a slight bias towards predictions, as Stage-1 already teaches rationale generation.

\paragraph{Transition Phase:} Exposure bias arises in autoregressive rationale generation from the train-test mismatch: training conditions explanations on gold labels, but inference must use model predictions, risking error propagation. During $w\leq t<w+m$, we gradually increase the fraction of predicted labels used for conditioning from $0$ to $0.9$. This transition period is to systematically address exposure bias by training the model to learn coherent rationale generation conditioned on its own (potentially imperfect) predictions, bridging the gap between training and inference conditions.

\smallskip{}

The explanation loss is:
\[
\mathcal{L}_{\text{expl}} = -\log P_\theta(r^* \mid \text{prompt})
\]
where prompt = $\texttt{given label:} \tilde{y} \texttt{, explain:} x$. \( P_\theta(r^* \mid \cdot) \) denotes the autoregressive probability of the full rationale token sequence.

\smallskip{}

 The total loss combines both objectives with adaptive weighting:
\[
\mathcal{L}_{\text{total}} = \alpha_{t} \mathcal{L}_{\text{pred}} + (1 - \alpha_{t}) \mathcal{L}_{\text{expl}}
\]

\subsection{Inference}
During inference, we first predict the label via greedy decoding:
\[\hat{y} = \arg\max_{y} P_\theta(y \mid \texttt{predict:}  x)\]
 and then generate the rationale, via greedy decoding, conditioned on that prediction:
\[
\hat{r} = \arg\max_{r} P_\theta(r \mid \text{prompt}),
\]
where prompt = $\texttt{given label:} \hat{y} \texttt{, explain:} x$.

\section{Experiments}

Here we introduce the datasets and provide implementation details, followed by the experimental results.
\subsection{Tasks and Datasets}
We evaluate Reason2Decide on one proprietary clinical decision-making task (triage notes), and two public biomedical QA benchmarks (PubMedQA, BioASQ). As models we use T5-Small/Base/Large and zero-shot LLMs as non-fine-tuned references. 
\subsubsection{Clinical Triage Dataset}
The Clinical Triage Dataset is derived from Alberta Health Link 811, a provincial telephone-based health advice service operated by Alberta Health Services (AHS), Canada. The dataset contains triage notes authored by nurses during telephone consultations with patients, along with the recommended care disposition and a rationale explaining the decision. 

\textbf{Task:} Given a nurse triage note, predict a disposition (care pathway) and generate a rationale for that recommendation.
The label space consists of 12 classes ranging from low (Home Care) to high urgency (e.g., "Go to Emergency Department (ED) now").\footnote{A complete list of the 12 disposition classes is provided in the Appendix, Subsection~\ref{subsec:definitions}}

\textbf{Rationale sources:} We use three versions of rationales to assess source robustness:
\begin{itemize}
    \item Nurse-authored rationales (original)
    \item Nurse post-processed rationales (lightly edited by an LLM)
    \item LLM-generated rationales (full generation using an LLM)
\end{itemize}
Table \ref{tab:rationale_sources_samples} shows sample rationales of each type. The post-processed version is there because the original version consists of atomic facts and is not always grammatically correct. We use\footnote{All prompts used are provided in the Appendix, Subsection ~\ref{subsec:prompts}.} a Qwen-3 8B \cite{qwen3technicalreport} model to convert them into grammatically correct full sentences without adding new information. For LLM-generated notes, we provide the model the triage note and disposition and ask it to explain why the disposition was chosen. For experiments on the whole dataset, we used the LLM-generated rationales.

\setlength{\arrayrulewidth}{0.1pt}


\newcolumntype{Y}{>{\raggedright\arraybackslash}X}

\begin{table}[h]
\footnotesize
\setlength{\tabcolsep}{4pt}
\renewcommand{\arraystretch}{1.18}
\begin{tabularx}{\linewidth}{YYY}
\toprule
\textbf{Nurse-authored} & \textbf{Post-processed} & \textbf{LLM-generated}\\
\midrule
{[}1{]} Earache AND {[}2{]} MODERATE pain OR SEVERE pain inadequately treated per guideline advice - yes
&
The patient has an earache with moderate or severe pain inadequately treated according to guideline advice.
&
Persistent severe ear pain without fever or infection signs requires timely medical evaluation to prevent complications and ensure appropriate treatment.
\\
\midrule
{[}1{]} MILD-MODERATE pain AND {[}2{]} constant AND {[}3{]} present > 2 hours
&
The patient has mild to moderate pain that is constant and has been present for more than two hours.
&
Urgent evaluation needed due to persistent abdominal pain, bowel changes, and recent confusion, to rule out serious conditions and ensure appropriate treatment.
\\
\bottomrule
\end{tabularx}
\caption{Sample rationale variants.}
\label{tab:rationale_sources_samples}
\end{table}

\textbf{Data split:} The train, validation and test sets consist of roughly 171k, 21k and 9.7k samples, respectively.

\subsubsection{PubMedQA}
\textbf{Task:} Biomedical question answering with Yes/No/Maybe labels \cite{jin-etal-2019-pubmedqa}. We concatenate the question and context as the model input as triage note replacement.

\textbf{Rationales:}  We use the dataset’s \texttt{long\_answer} field as the gold-standard rationale.

\textbf{Data split:} We draw 100k stratified random samples (by labels) using a fixed seed from pqa\_artificial (the dataset artificially generated), and create a 70k/20k/10k train/validation/test split. We then augment the test set with the 1k samples in pqa\_labeled, as they are human annotated data, resulting in final splits of 70k (train), 20k(validation), and 11k (test: 10k pqa\_artificial + 1k pqa\_labeled).

\subsubsection{BioASQ (Task B, Yes/No)}
\paragraph{Task:} Biomedical question answering dataset from the BioASQ 13 challenge \cite{nentidis2025overviewbioasq2025thirteenth}. For our experiments, we focus on \texttt{yes/no} questions (binary classification).

\paragraph{Rationales:} We extract and concatenate the \texttt{snippets} to serve as rationales for each question. Because the concatenated snippets can be long, we first summarize them with a Qwen-3-8B model and use the summary as the gold rationale.

\paragraph{Data split:} The official training dataset consists of 1,459 \texttt{yes/no} questions. We randomly select 250 questions for our validation set, and 1,209 questions for the train set. For the test set, we use the concatenation of the four official test batches, yielding 82 questions. 

\subsection{Implementation Details}
We implement\footnote{Source code available at: \url{https://github.com/quamranhasan/Reason2Decide}} Reason2Decide using T5 Small (77M), Base (250M), Large (800M) architectures using A100 X 4 GPUs. Training uses AdamW optimizer \cite{loshchilov2019decoupledweightdecayregularization} with learning rate $5\times10^{-5}$, max input length = 1024, and effective batch size = 64. Model-specific configuration is provided in Table \ref{tab:train-config}. 

For Stage-1, we monitor validation loss and early stop with patience of 3 validation evaluations. For Stage-2, we use delayed early stopping, with a patience of 5, that activates only after the scheduled sampling phase completes. This is to ensure the model fully benefits from our warm-up and scheduled sampling recipe. 

During validation, we evaluate only the prediction task using F1-score, treating rationales as an auxiliary training signal. The best model checkpoint (highest F1-score) is loaded at the end of training. The models are trained using publicly available packages from https://github.com/huggingface/transformers.

\begin{table}[h]
\centering
\footnotesize
\setlength{\tabcolsep}{8pt}
\resizebox{0.4\textwidth}{!}{
\begin{tabular}{lcccc}
\toprule
\textbf{Model}& \textbf{Per-GPU Batch} & \textbf{Grad Accum}\\
\midrule
T5-Small & $16$ & $1$ &\\
T5-Base  & $4$  & $4$ &\\
T5-Large & $2$  & $8$ &\\
\bottomrule
\end{tabular}
}
\caption{Training configuration by model size. Effective batch size is Per-GPU Batch $\times$ \#GPUs $\times$ Grad Accum.} 
\label{tab:train-config}
\end{table}

\subsection{Baselines}
We compare Reason2Decide against the following:

\begin{itemize}
    \item \textbf{Standard fine-tuning (SFT):} Single-task label prediction without rationales \cite{howard-ruder-2018-universal}; model selection by validation Macro-F1.
    \item \textbf{Distilling Step-by-Step (DSS):} Multi-task training following \cite{hsieh-etal-2023-distilling}. We run two selection criteria:
    \begin{itemize}
        \item \textbf{DSS-Loss:} Model selection by validation loss (as in \citet{hsieh-etal-2023-distilling}).
        \item \textbf{DSS-F1:} Model selection by validation Macro-F1 (to align with our selection protocol).
    \end{itemize}
    Implementation follows the authors’ release. 
    \item \textbf{Zero-shot LLMs:} Zero-shot baselines without task-specific fine-tuning, using open-source models including Qwen-3-8B and Qwen-3-32B~\cite{qwen3technicalreport}, as well as Llama-3.1-Aloe-Beta-8B \cite{garcia2025aloe}.
\end{itemize}

\paragraph{Protocol:} For all fine-tuned baselines, we use the same optimizer, learning rate, and effective batch size as in our method; per-device batch sizes are set following Table~\ref{tab:train-config}. Early stopping is used with patience of 5 validation evaluations. Model selection is by validation Macro-F1 unless otherwise stated (DSS-Loss). We report means over three seeds/runs. We use greedy decoding to ensure run-to-run determinism.


\subsection{Evaluation}
We provide a comprehensive evaluation across predictive performance and rationale quality:

\paragraph{Predictive Performance:} We report Accuracy and Macro-F1 on the discrete label space (dispositions for triage; Yes/No/Maybe for PubMedQA; Yes/No for BioASQ). Scores are computed on held-out test sets. Due to the high class imbalance in the Triage Dataset (major and minor classes have $33,633$ and $239$ samples respectively in the training set), we primarily rely on Macro-F1.

\paragraph{Rationale Fidelity:} We assess explanation quality with:
\begin{itemize}
    \item \textbf{BERTScore} \cite{zhang2020bertscoreevaluatingtextgeneration} (F1 variant): Semantic similarity between generated and gold rationales. 
    \item \textbf{BLEU} \cite{papineni-etal-2002-bleu}: N-gram overlap for surface-level quality.
    \item \textbf{LLM-as-a-Judge}: Following recent work demonstrating strong alignment between LLM and human evaluation \cite{zheng2023judgingllmasajudgemtbenchchatbot, niu-etal-2025-knowledge}, we employ Qwen-3-8B for expert-style assessment on random 2k samples for the Triage Dataset and PubMedQA, and the whole test set for BioASQ (82 samples). We evaluate three defined metrics:
    
\textbf{Correctness}: 5-point scale for clinical alignment between the rationales and predictions. The motivation is ``If the predicted disposition is \texttt{Go to L\&D now}, does the generated rationale justify the decision - not \texttt{homecare?}" To ensure consistent evaluation, we generated standardized disposition definitions using prompt-based refinement with a Qwen-3-32B model. 

\textbf{Coverage}: 5-point scale that evaluates information retention. Scores measure how much relevant clinical information from the generated rationale is in the original triage note.

\textbf{Overlap}: 5-point scale for semantic content preservation between the gold and generated rationale.

\end{itemize}

For the triage dataset, we report all three LLM-as-a-Judge metrics. For PubMedQA and BioASQ, we only report  \texttt{overlap}, as the others do not strictly align with QA datasets, due to missing triage notes and simple \texttt{(yes/no/maybe)}labels, respectively.

\subsection{Results}

\begin{table*}[h]
\centering
\footnotesize
\resizebox{0.7\textwidth}{!}{
\begin{tabular}{llcccc}
\toprule
\textbf{Rationale Source} & \textbf{Method} & \textbf{Macro F1} & \textbf{BERTSc.} & \textbf{BLEU} & \textbf{Overlap} \\
\midrule
 & SFT & $48.78 \pm 0.44$ & - & - & - \\
 \midrule
 \multirow{3}{*}{LLM-generated} 
 & DSS-F1 & $49.69 \pm 0.58$ & $90.68 \pm 0.12$ & $15.00 \pm 0.44$ & $2.43$ \\
 & DSS-Loss & $47.27 \pm 0.40$ & $90.38 \pm 0.05$ & $13.72 \pm 0.17$ & $2.37$ \\
 & \textbf{Reason2Decide} & $\mathbf{50.06 \pm 0.14}$ & $\mathbf{91.02 \pm 0.02}$ & $\mathbf{16.46 \pm 0.06}$ & $\mathbf{2.50}$ \\
\midrule
\multirow{3}{*}{Nurse-authored}
 & DSS-F1 & $48.01 \pm 0.39$ & $87.67 \pm 0.06$ & $24.88 \pm 0.40$ & $2.51$ \\
 & DSS-Loss & $46.76 \pm 1.02$ & $87.51 \pm 0.07$ & $23.87 \pm 0.39$ & $2.47$ \\
 & \textbf{Reason2Decide} & $\mathbf{50.76 \pm 1.01}$ & $\mathbf{88.06 \pm 0.23}$ & $\mathbf{26.62 \pm 1.44}$ & $\mathbf{2.62}$ \\
\midrule
\multirow{3}{*}{Post-processed}
 & DSS-F1 & $47.48 \pm 2.11$ & $90.15 \pm 0.13$ & $22.47 \pm 0.85$ & $2.54$ \\
 & DSS-Loss & $45.05 \pm 3.07$ & $89.89 \pm 0.03$ & $20.68 \pm 0.23$ & $2.48$ \\
 & \textbf{Reason2Decide} & $\mathbf{51.14 \pm 0.44}$ & $\mathbf{90.46 \pm 0.07}$ & $\mathbf{24.03 \pm 0.55}$ & $\mathbf{2.64}$ \\
\bottomrule
\end{tabular}
}
\caption{Rationale Source Robustness on Clinical Triage (T5-Small on 12\% of Clinical Triage Dataset)}
\label{tab:source-robustness}
\end{table*}

\begin{table*}[h!]
\centering
\resizebox{0.8\textwidth}{!}{
\begin{tabular}{lllcccc}
\toprule
\textbf{Dataset} & \textbf{Model} & \textbf{Method} & \textbf{Macro F1} & \textbf{Accuracy}& \textbf{BERTSc.} & \textbf{BLEU} \\
\midrule

\multirow{16}{*}{Clinical Triage}
& \multirow{4}{*}{T5-Small} & SFT & $52.69 \pm 0.86$ & - & - & - \\

& & DSS-Loss & $52.73 \pm 0.99$ & - & $91.14 \pm 0.09$ & $17.37 \pm 0.45$ \\

& & DSS-F1 & $52.09 \pm 0.50$ & - & $90.99 \pm 0.06$ & $16.72 \pm 0.33$ \\

& & \textbf{{Reason2Decide}} & $\mathbf{55.88 \pm 0.01}$ & - & ${91.52 \pm 0.06}$ & ${19.33 \pm 0.29}$ \\
\cmidrule{2-7}

& \multirow{5}{*}{T5-Base} & SFT & $57.10 \pm 1.15$ & - & - & - \\
& & DSS-Loss & $53.26 \pm 0.89$ & - & $91.78 \pm 0.04$ & $20.86 \pm 0.27$ \\
& & DSS-F1 & $54.53 \pm 2.63$ & - & $91.56 \pm 0.32$ & $19.50 \pm 1.70$ \\
& & \textbf{{Reason2Decide}} & $\mathbf{59.92 \pm 0.42}$ & - & ${92.09 \pm 0.02}$ & ${22.74 \pm 0.06}$ \\
\cmidrule{2-7}

& \multirow{4}{*}{T5-Large} & SFT & $56.85 \pm 7.21$ & - & - & - \\
& & DSS-Loss & $58.09 \pm 0.77$ & -  & $92.08 \pm 0.05$ & $22.73 \pm 0.17$ \\
& & DSS-F1 & $59.43 \pm 1.07$ & -  & ${92.20 \pm 0.03}$ & $23.56 \pm 0.09$ \\
& & \underline{\textbf{Reason2Decide}} & \underline{$\mathbf{60.58 \pm 0.46}$} & -  & ${92.30 \pm 0.03}$ & ${24.13 \pm 0.19}$ \\

\cmidrule{2-7}

& Aloe-8B & \multirow{3}{*}{Zero-Shot} & $24.73 \pm 0.00$ & - & - & - \\
& Qwen-3-8B &  & $23.28 \pm 0.00$ & - & - & - \\
& Qwen-3-32B &  & $28.03 \pm 0.00$ & - & - & - \\

\midrule

\multirow{15}{*}{PubMedQA}
& \multirow{4}{*}{T5-Small} & SFT & $52.18 \pm 0.07$ & $91.87 \pm 0.34$ & - & - \\

& & DSS-F1 & $53.63 \pm 0.57$ & $93.12 \pm 0.16$ & $89.14 \pm 0.01$ & $6.36 \pm 0.06$ \\

& & DSS-Loss & $53.63 \pm 0.07$ & $93.14 \pm 0.08$ & $89.11 \pm 0.05$ & $6.27 \pm 0.07$ \\

& & \textbf{{Reason2Decide}} & $\mathbf{55.02 \pm 0.16}$ & ${93.51 \pm 0.14}$ & ${89.30 \pm 0.03}$ & ${6.76 \pm 0.15}$ \\

\cmidrule{2-7}

& \multirow{4}{*}{T5-Base} & SFT & $57.31 \pm 0.16$ & $94.58 \pm 0.11$ & - & - \\

& & \textbf{DSS-F1} & $\mathbf{58.16\pm 0.11}$ & $95.00 \pm 0.03$ & $89.40 \pm 0.04$ & $6.93 \pm 0.13$ \\

& & DSS-Loss & $57.69 \pm 0.25$ & $94.79 \pm 0.14$ & $89.40 \pm 0.01$ & $6.82 \pm 0.10$ \\

& & {{Reason2Decide}} & ${58.10 \pm 0.22}$ & ${95.04 \pm 0.09}$ & ${89.51 \pm 0.03}$ & ${7.17 \pm 0.04}$ \\

\cmidrule{2-7}

& \multirow{4}{*}{T5-Large} & SFT & $59.60 \pm 0.26$ & $95.72 \pm 0.17$ & - & - \\

& & DSS-F1 & $59.92 \pm 0.21$ & $95.90 \pm 0.06$  & ${89.62 \pm 0.04}$ & $7.41 \pm 0.12$ \\

& & DSS-Loss & $59.74 \pm 0.26$ & $95.70 \pm 0.28$ & $89.61 \pm 0.03$ & $7.08 \pm 0.09$ \\

& & \underline{\textbf{Reason2Decide}} & $\underline{\mathbf{60.28 \pm 0.05}}$ & ${96.05 \pm 0.01}$ & $89.60 \pm 0.05$ & ${7.64 \pm 0.15}$ \\

\cmidrule{2-7}

& Aloe-8B & \multirow{3}{*}{Zero-Shot}  & $45.33 \pm 0.00$ & $95.25 \pm 0.00$ & - &  - \\
& Qwen-3-8B & & $52.85 \pm 0.00$ & $87.89 \pm 0.00$ & -  & - \\
& Qwen-3-32B & & $58.33 \pm 0.00$ & $90.83 \pm 0.00$ & -  & - \\
\midrule

\multirow{15}{*}{BioASQ}

& \multirow{4}{*}{T5-Small} & SFT & $64.81 \pm 2.09$ & $72.76 \pm 1.86$ & - & - \\

& & DSS-F1 & $66.02 \pm 1.19$ & $71.14 \pm 0.70$ & $84.16 \pm 0.43$ & $2.68 \pm 0.14$ \\

& & DSS-Loss & $40.47 \pm 2.14$ & $65.04 \pm 1.41$ & $82.80 \pm 0.39$ & $1.86 \pm 0.15$ \\

& & \textbf{{Reason2Decide}} & $\mathbf{67.57 \pm 2.26}$ & ${73.58 \pm 0.70}$ & ${85.14 \pm 0.18}$ & ${3.27 \pm 0.09}$ \\

\cmidrule{2-7}

& \multirow{4}{*}{T5-Base} & SFT & $53.70 \pm 12.29$ & $67.88 \pm 2.54$  & - & - \\

& & DSS-F1 & $66.28 \pm 0.54$ & $71.54 \pm 1.86$  & ${85.78 \pm 0.14}$ & ${3.85 \pm 0.02}$ \\

& & DSS-Loss & $59.99 \pm 13.58$ & $71.14 \pm 6.72$ & X & X \\

& & \textbf{{Reason2Decide}} & $\mathbf{68.02 \pm 2.19}$ & 
${73.98 \pm 1.41}$ & $85.64 \pm 0.30$ & $3.55 \pm 0.10$ \\
\cmidrule{2-7}

& \multirow{4}{*}{T5-Large} & \textbf{SFT} & $\mathbf{66.80 \pm 2.64}$ & $73.98 \pm 1.41$ & - & - \\

& & DSS-F1 & $65.73 \pm 2.15$ & $70.73 \pm 2.44$ & ${85.79 \pm 0.25}$ & ${3.43 \pm 0.29}$ \\

& & DSS-Loss & $44.55 \pm 4.19$ & $67.48 \pm 1.41$ & X & X \\

& & {{Reason2Decide}} & ${66.58 \pm 4.57}$ & ${73.17 \pm 2.44}$ & $85.80 \pm 0.11$ & $3.39 \pm 0.26$ \\
\cmidrule{2-7}

& \underline{\textbf{Aloe-8B}} & \multirow{3}{*}{Zero-Shot} &  \underline{$\mathbf{79.09 \pm 0.00}$} & $81.71 \pm 0.00$ & -& - \\
& Qwen-3-8B & & $50.32 \pm 0.00$ & $75.61 \pm 0.00$ & - & - \\
& Qwen-3-32B & & $50.69 \pm 0.00$ & $76.83 \pm 0.00$ & -  & - \\
\bottomrule
\end{tabular}
}
\caption{Performance comparison across datasets. Best F1 score per dataset  is \underline{\textbf{bold+underlined}}. Best fine-tuning strategy per model size is \textbf{{bold}}. Standard deviations are shown as $\pm$ values. Instances where rationale generation is non-applicable are shown as -. Instances where the model failed to generate a rationale are shown as X.}
\label{tab:combined_results}
\end{table*}

\subsubsection{Rationale Source Robustness}
We evaluate Reason2Decide's robustness to rationale source variations by training Stage-1 exclusively on LLM-generated rationales, and in Stage-2, using the matching rationale variant (LLM-generated, nurse-authored, post-processed). As shown in Table~\ref{tab:source-robustness}, Reason2Decide maintains consistent performance across rationale variants while outperforming all baselines on both prediction and rationale metrics, with +0.37 to +6.09 F1 over DSS variants. Although this experiment was conducted on only 12\% of the dataset, our method produces observable gains in data-scarce clinical situations. 

Despite training on LLM-generated rationales only during Stage-1, Reason2Decide shows 51\% lower F1 variation than DSS-F1 (1.08-point spread vs 2.21) and better rationale metrics. The F1 gains demonstrate that rationale pretraining provides transferable reasoning benefits that improve prediction, and not only explanation generation. This pattern suggests that LLM rationales may substitute for human-authored rationales during pretraining, thereby reducing reliance on costly human rationales. Our method demonstrates strong rationale source robustness, successfully adapting to different rationale styles during Stage-2 training, achieving best performance on post-processed rationales (51.14 F1) with consistent gains across rationale sources in BERTScore, BLEU, and LLM-as-a-Judge Overlap.

\subsubsection{Consistent Performance Across Model Sizes and Tasks}

From Table \ref{tab:combined_results}, across all three datasets, Reason2Decide shows improvement over other fine-tuned variants. For Clinical Triage, Reason2Decide outperforms every other baseline, across model sizes. 

Similar results are observed with PubMedQA, with Reason2Decide outperforming other variants with the Small and Large models. Despite slightly lagging behind DSS-F1 with the Base model, Reason2Decide outperforms DSS-F1 on BLEU and BERTScore.

BioASQ proved to be a challenging dataset for all fine-tuned variants, possibly due to its limited data size. Although Reason2Decide outperforms the baselines on the Small and Base models, SFT takes the win with the Large model.

Reason2Decide outperformed both Qwen models on all datasets. However, Aloe-8B scores the highest on BioASQ. This can be attributed to its  pretraining on biomedical literature, which closely matches the domain and format of BioASQ, but differs from the conversational and reasoning-oriented style of the triage and PubMedQA datasets.

While T5-Large (800M) is $40\times$ smaller than the 32B Qwen model and is not a foundation model pretrained on medical corpora, it outperformed Qwen-3-32B on all three datasets and Aloe-8B on 2/3 datasets, narrowing the performance gap between LLMs and smaller language models.

\subsubsection{Rationale Analysis}

\textbf{BLEU and BERTScore:} As shown in Table~\ref{tab:combined_results}, on Clinical Triage, for all three model sizes, Reason2Decide outperforms DSS variants in both metrics. On PubMedQA and BioASQ, we observe similar trends with Reason2Decide scoring competitively with DSS variants. 

\textbf{LLM-as-a-Judge Evaluation:} To assess the quality of generated rationales, we evaluate three metrics as defined earlier. These were scored on a $(1-5)$ Likert scale. From Table \ref{tab:rationale_quality}, Reason2Decide scores the highest in all metrics with the Small and Base model variants. With the Large model, the \texttt{overlap} and \texttt{coverage} scores are very close, with our proposed method scoring comparatively higher in \texttt{correctness}. 

A sample prediction and rationale is provided in Table \ref{tab:qualitative}.
\begin{table}[h]
\centering
\small
\resizebox{0.48\textwidth}{!}{
\begin{tabular}{lcccc}
\toprule
\textbf{Model Size} & \textbf{Method} & \textbf{Coverage} & \textbf{Correctness} & \textbf{Overlap} \\
\midrule
\multirow{3}{*}{T5-Small} 
  & DSS-Loss       & $4.52$ & $4.02$ & $2.50$ \\
  & DSS-F1         & $4.50$ & $3.96$ & $2.48$ \\
  & Reason2Decide   & $\mathbf{4.62}$ & $\mathbf{4.26}$ & $\mathbf{2.58}$ \\
\midrule
\multirow{3}{*}{T5-Base} 
  & DSS-Loss       & $4.74$ & $4.09$ & $2.57$ \\
  & DSS-F1         & $4.71$ & $4.05$ & $2.55$ \\
  & Reason2Decide   & $\mathbf{4.76}$ & $\mathbf{4.37}$ & $\mathbf{2.61}$ \\
\midrule
\multirow{3}{*}{T5-Large} 
  & DSS-Loss       & $4.78$ & $4.13$ & $2.60$ \\
  & DSS-F1         & $\mathbf{4.81}$ & $4.19$ & $2.61$ \\
  & Reason2Decide   & $4.80$ & $\mathbf{4.43}$ & $\mathbf{2.63}$ \\
\bottomrule
\end{tabular}
}
\caption{LLM-as-a-Judge evaluation metrics.}
\label{tab:rationale_quality}
\end{table}

\begin{table}[h]
\centering
\resizebox{0.48\textwidth}{!}{
\begin{tabular}{p{0.15\textwidth}p{0.65\textwidth}}
\hline
\textbf{Method} & \textbf{Rationale and Prediction} \\
\hline
\textbf{Gold Standard} & \textit{Rationale:} Infant has repeated episodes of gasping for air, indicating possible \underline{serious breathing difficulty} requiring immediate medical evaluation. \\
& \textit{Prediction:} Go to ED Now \\
\hline
\textbf{Reason2Decide} & \textit{Rationale:} Infant has repeated episodes of gasping for air, indicating possible \underline{serious breathing difficulty} requiring immediate medical evaluation. \\
& \textit{Prediction:} Go to ED Now \\
\hline
\textbf{DSS-F1} & \textit{Rationale:} Infant's frequent gasping episodes with fever require immediate evaluation to rule out serious respiratory issues despite normal appearance and no worsening symptoms. \\
& \textit{Prediction:} See Physician or PCP within 24 Hours \\
\hline
\textbf{DSS-Loss} & \textit{Rationale:} Infant's frequent gasping episodes with \underline{normal breathing} and no fever require prompt medical evaluation to rule out serious respiratory issues. \\
& \textit{Prediction:} Go to ED Now \\
\hline
\end{tabular}
}
\caption{Sample rationale and prediction. Reason2Decide produced gold-aligned rationale and prediction; DSS-F1 gave contradicting rationale and prediction; DSS-Loss made correct prediction but with inaccurate rationale.}
\label{tab:qualitative}
\end{table}

\subsection{Ablation Study}
We performed ablations to assess the contribution of each component in our Reason2Decide framework. Table \ref{tab:ablation} shows that removing Stage-1 consistently hurts performance across datasets, confirming that explanation-focused pretraining improves downstream decision-making. Stage-2 is essential for predictions, because the Stage-1 model only learns to generate explanations. 

The primary purpose of scheduled sampling is rationale alignment via exposure-bias mitigation. This is reflected with its equal or higher BERT and BLEU scores across both datasets. Additionally, it improves prediction on complex, multi-class datasets like Clinical Triage (59.92 vs. 57.28 F1), suggesting that training with realistic inference conditions could provide a greater benefit for challenging prediction tasks. 

Including warm-up steps slightly improved performance on Clinical Triage across all metrics. Removing warm-up steps on BioASQ caused a substantial F1 drop despite small BERT/BLEU gains, indicating the importance of gradual adaptation from the single-task to the multi-task objective in Stage-2.

\begin{table}[h]
\centering
\resizebox{0.45\textwidth}{!}{
\begin{tabular}{lcccc}
\toprule
\textbf{Method} & \textbf{Accuracy} & \textbf{F1} & \textbf{BERT} & \textbf{BLEU} \\
\midrule
\multicolumn{5}{c}{\textbf{Clinical Triage}} \\
\midrule
\textit{w/o} Stage-1 & \textemdash & $58.25$ & $91.65$ & $19.95$ \\
\textit{w/o} Stage-2 & \textemdash & $0.00$  & $91.78$ & $20.72$ \\
\textit{w/o} Scheduled sampling     & \textemdash & $57.28$ & $\mathbf{92.09}$ & $22.59$ \\
\textit{w/o} Warm-up steps & \textemdash & $59.51$ & $92.07$ & $22.64$ \\
Reason2Decide & \textemdash & $\mathbf{59.92}$ & $\mathbf{92.09}$ & $\mathbf{22.74}$ \\
\midrule
\multicolumn{5}{c}{\textbf{BioASQ}} \\
\midrule
\textit{w/o} Stage-1 & $71.95$ & $65.99$ & $85.55$ & $3.32$ \\
\textit{w/o} Stage-2 & $0.00$  & $0.00$  & $84.17$ & $2.95$ \\
\textit{w/o} Scheduled sampling     & $72.76$ & $\mathbf{68.06}$ & $84.15$ & $3.07$ \\
\textit{w/o} Warm-up steps & $69.10$ & $61.27$ & $\mathbf{85.68}$ & $\mathbf{3.71}$ \\
Reason2Decide & $\mathbf{73.98}$ & $68.02$ & $85.64$ & $3.55$ \\
\bottomrule
\end{tabular}
}
\caption{Ablation results on Clinical Triage and BioASQ (Base model).}
\label{tab:ablation}
\end{table}

\section{Conclusion and Future Work}
We have introduced Reason2Decide, a two-stage training framework for LLMs designed to enhance decision quality and interpretability in clinical NLP tasks. Through this framework, the model learns to generate  rationales that align with its predictions. Experiments on nurse triage and biomedical QA datasets show that Reason2Decide outperforms other fine-tuning variants in both prediction and explanation metrics. With our low cross-source F1 variation, we show that synthetic rationales can substitute for costly human rationales.

In future work, our aim is to extend this approach to broader clinical and biomedical subdomains, to assess its generalizability. We also want to investigate the effect of integrating rule-based or CoT-style rationales. In addition, our goal is to add human evaluation for better assessing the rationales. 

\section{Limitations}
While Reason2Decide demonstrates strong performance over other fine-tuning variants, certain limitations remain. Firstly, the datasets used in this work fall under nurse triage and biomedical QA. To effectively assess Reason2Decide, it should be extended to other subdomains in clinical NLP. Secondly, the LLM-generated rationales were not rule-based. It would be valuable to see the impact of using CoT-style rationales during Stage-1 on predictions. Moreover, our two-stage training framework requires increased computational resources during training. However, inference is fully end-to-end from the final Stage-2 checkpoint alone: it both predicts the label and generates self-conditioned explanations. Our approach involved several hyperparameters that were chosen heuristically. A more granular hyperparameter optimization strategy may lead to Reason2Decide's further improved performance. Finally, our rationale evaluation lacks human oversight. Although LLM-as-a-Judge methods are becoming increasingly adopted, they may be imperfect. For high-stakes domains like clinical NLP, human verification remains essential. 

\section{Ethics Statement}
During this research, we ensured to follow ethical guidelines for clinical NLP. The proprietary dataset used was de-identified. No personally identifiable information was accessible to the models or researchers during training/evaluation. Only open-source models were used, and were run on local machines. All clinical predictions and rationales generated by Reason2Decide should be treated as a decision support tool requiring human verification and not a replacement for human decisions. 

\section{Acknowledgements}
We gratefully acknowledge Alberta Health Services (AHS) for providing access to the Alberta Health Link 811 dataset for research purposes.
This work was supported by the Natural Sciences and Engineering Research Council of Canada (NSERC) Collaborative Research and Training Experience (CREATE) program “From Data to Decision” (FD2D). This research was also supported by the Alberta Machine Intelligence Institute (Amii), NSERC (including grants DGECR-2022-00369 and RGPIN-2022-0346), and Alberta Innovates.

\section{Bibliographical References}\label{sec:reference}

\bibliographystyle{lrec2026-natbib}
\bibliography{lrec2026-example}

\label{lr:ref}
\bibliographystylelanguageresource{lrec2026-natbib}
\bibliographylanguageresource{languageresource}

\section{Appendix}
\raggedright
\setcounter{table}{0}
\renewcommand{\thetable}{A\arabic{table}}

\setcounter{figure}{0}
\renewcommand{\thefigure}{S\arabic{figure}} 

\subsection{Prompts Used}
\label{subsec:prompts}
\paragraph{Prompt for Post-Processing Nurse-Authored Rationales:}
\label{subsec:postprocess_prompt}

\begin{center}
\fbox{\begin{minipage}{0.95\columnwidth}
\raggedright
You are a helpful assistant who expands brief medical notes into full, grammatically correct sentences using fewer than 20 words. Do not add new information.\\

Convert this to a sentence without adding new information: [RATIONALE]
\end{minipage}}
\end{center}
\vspace{0.5em}
This created the nurse post-processed rationale variant by replacing [RATIONALE] with each original nurse note.

\paragraph{Prompt for Generating Standardized Disposition Definitions:}
\label{subsec:definition_prompt}

\begin{center}
\fbox{\begin{minipage}{0.95\columnwidth}
\raggedright
You are building a clinical definition for the disposition: [DISPOSITION\_NAME].

\vspace{1em}

Current working definition: 

\vspace{1em}
[CURRENT\_DEFINITION]
\vspace{1em}

You are given new examples of triage notes and rationales for this disposition. 
Use them to refine, expand, or correct the working definition. 
Keep the definition concise but clinically accurate.

\vspace{1em}

Triage Notes and Rationales:
[EXAMPLES]

\vspace{1em}

Update the definition:

- Incorporate any new key symptoms, criteria, or thresholds.

- Remove incorrect parts.

- Keep it as clear and specific as possible.

Output only the revised definition text, nothing else.
\end{minipage}}
\end{center}
\vspace{0.5em}
Where [DISPOSITION\_NAME], [CURRENT\_DEFINITION], and [EXAMPLES] were replaced with actual data. This process generated the standardized definitions used for consistent LLM-as-a-Judge scoring.

\paragraph{Prompt for Zero-Shot Disposition Classification:}
\label{subsec:zeroshot_prompt}

\begin{center}
\fbox{\begin{minipage}{0.95\columnwidth}
\raggedright
Issue: [ISSUE\_ASSESSMENT]

Dispositions: [CLASSES\_TEXT]

\vspace{1em}

Classify the healthcare issue into one of the dispositions above.
Return your answer in the following **strict** format:

Class: [chosen digit]

\vspace{1em}

Do not ask for more information, and do not provide any general statements. Only respond with the digit.
\end{minipage}}
\end{center}
\vspace{0.5em}
Where [ISSUE\_ASSESSMENT] contained the triage note text and [CLASSES\_TEXT] listed the 12 disposition options with their numerical identifiers. This prompt was used for zero-shot LLM evaluation.

\paragraph{Prompt for Summarizing BioASQ Snippets:}
\label{subsec:summary_prompt}
\begin{center}
\fbox{\begin{minipage}{0.95\columnwidth}
\raggedright
Please summarize the following medical rationale in 100 words or less. 
Focus on the key points and main conclusions. Keep it concise and informative.

\vspace{1em}

Rationale to summarize: [RATIONALE]

Summary:
\end{minipage}}
\end{center}

Where [RATIONALE] was replaced with the concatenated text snippets from BioASQ. This created concise gold-standard rationales for BioASQ experiments.

\paragraph{Prompt for LLM-as-a-Judge Correctness Scoring:}
\label{subsec:correctness_prompt}
\begin{center}
\fbox{\begin{minipage}{0.95\columnwidth}
\raggedright
You are an expert clinical trainer for telephone triage nursing.

[DEFINITIONS]

\vspace{1em}

Task:
Given the following rationale and disposition, score the alignment on a scale of 1 to 5, where:

5 - Excellent Alignment

4 - Good Alignment

3 - Moderate Alignment

2 - Poor Alignment

1 - Very Poor Alignment

\vspace{1em}

Rationale: [RATIONALE]

Disposition: [DISPOSITION]

Output exactly one number (1, 2, 3, 4, or 5) with no other text.
\end{minipage}}
\end{center}
\vspace{0.5em}
Where [DEFINITIONS] was replaced with the standardized disposition definitions following Subsection~\ref{subsec:definitions}, [RATIONALE] with the generated rationale, and [DISPOSITION] with the predicted disposition.

\paragraph{Prompt for LLM-as-a-Judge Coverage Scoring:}
\label{subsec:coverage_prompt}

\begin{center}
\fbox{\begin{minipage}{0.95\columnwidth}
You are a medical text comparison assistant.

Task: Compare the triage note with the text provided. 
Determine how much of the information in the text is also present in the triage note.

\vspace{1em}

Output only one number based on this scale:

5 = The note contains everything mentioned in the text.

4 = The note contains most things mentioned in the text.

3 = The note contains some but not most things mentioned in the text.

2 = The note contains almost nothing mentioned in the text.

1 = The note contains nothing mentioned in the text.

\vspace{1em}

Triage Note:
[NOTE]

Text:
[TEXT]

Output only the number:
\end{minipage}}
\end{center}

Where [NOTE] was replaced with the original triage note and [TEXT] with the generated rationale.

\paragraph{Prompt for LLM-as-a-Judge Overlap Scoring:}
\label{subsec:overlap_prompt}

\begin{center}
\fbox{\begin{minipage}{0.95\columnwidth}
\raggedright
You are a clinical text comparison assistant.

Task: Compare the nurse's original rationale with the model's predicted rationale and rate how much of the nurse's rationale is present in the predicted rationale.
\vspace{1em}
Judge semantic content, not wording. Paraphrases/synonyms count as overlap. Ignore pleasantries and generic instructions. Contradictions reduce overlap.

Output exactly one number per line for each pair, with no other text:

1 = No overlap

2 = Almost no overlap

3 = Some but not most overlap

4 = Most overlap

5 = Complete overlap
\vspace{1em}

Nurse Rationale:
[NURSE\_RATIONALE]

Predicted Rationale:

[PREDICTED\_RATIONALE]

Output only the number:
\end{minipage}}
\end{center}

\vspace{0.5em}
Where [NURSE\_RATIONALE] was the gold-standard nurse rationale and [PREDICTED\_RATIONALE] was the model-generated rationale.

\subsection{Clinical Disposition Definitions}
\label{subsec:definitions}
\textbf{Home Care: } Patients suitable for Home Care disposition have mild to moderate, stable, and non-progressive symptoms without signs of immediate or severe complications requiring emergency intervention. They exhibit no respiratory distress, hemodynamic instability, severe pain unresponsive to treatment, significant bleeding, high fever with systemic symptoms, spreading infection, altered mental status, neurological deficits, or other urgent clinical concerns. Their condition allows for safe management, symptom relief, and observation at home with appropriate follow-up. This includes stable minor injuries, controlled localized infections, mild chronic condition issues, and non-urgent questions or concerns. Patients and caregivers should be advised to seek urgent care if symptoms worsen or new concerning signs develop.

\textbf{See Physician or PCP within 3 days: }Patients with mild to moderate, stable but persistent or worsening symptoms that do not pose an immediate threat to life or function, and who require timely medical evaluation to prevent complications. This includes conditions without severe pain, respiratory distress, hemodynamic instability, significant neurological deficits, acute infection, or other urgent signs. The disposition excludes any patients exhibiting signs of severe or rapidly progressing illness, acute psychiatric emergencies, or other urgent conditions necessitating immediate or emergency care.

\textbf{Call Pharmacist within 24 Hours: }This disposition is used when a caller has non-emergency medication-related questions or concerns that require timely pharmacist expertise within 24 hours to ensure safe, effective, and appropriate medication use. It applies when immediate urgent care is not needed but professional assessment is necessary to clarify dosing, manage side effects, verify medication information, address potential interactions, support adherence, assist with medication access, or provide guidance on special populations and circumstances. This ensures optimized therapy, prevention of harm, and informed patient decisions without delay.

\textbf{See Physician or PCP within 4 Hours (or PCP triage): }Patients with new or worsening moderate to severe symptoms, signs of infection, or conditions at risk of rapid deterioration that require timely clinical evaluation within hours to prevent complications. This includes significant pain unrelieved by initial treatment, progressive neurological symptoms, signs of systemic infection, post-procedural complications, unstable vital signs, moderate to severe respiratory, abdominal, or urinary symptoms, pregnancy-related concerns, dehydration, metabolic instability, mental health deterioration, and other clinical presentations indicating potential for rapid decline. The disposition ensures prompt physician assessment to guide urgent management and prevent adverse outcomes.

\textbf{Call EMS 911 Now: }Initiate immediate emergency medical services activation for any patient presenting with signs or symptoms of a potentially life-threatening condition. This includes acute neurological deficits, severe chest pain or cardiac symptoms, significant respiratory distress, signs of anaphylaxis, major trauma or uncontrolled bleeding, severe abdominal or back pain with systemic symptoms, acute deterioration in patients with serious underlying conditions, active suicidal intent with risk of harm, critical illness in infants or young children, shock or imminent collapse, severe infection or sepsis, altered mental status or unresponsiveness, and any situation posing an immediate threat to life, safety, or vital functions. Prompt EMS activation is essential whenever there is concern for compromised airway, breathing, circulation, neurological status, or urgent mental health crisis requiring rapid intervention.

\textbf{Go to ED Now: }Immediate emergency department evaluation is required for patients presenting with sudden, severe, or rapidly worsening symptoms that pose an immediate risk to life, limb, or function. This includes active uncontrolled bleeding, significant head or neck injuries, severe respiratory distress or airway compromise, acute neurological deficits, severe or persistent pain suggestive of surgical or obstetric emergencies, signs of shock or altered consciousness, severe allergic reactions, prolonged seizures, suspected serious infections, severe metabolic disturbances, and any other critical conditions requiring urgent assessment and intervention.

\textbf{See Physician or PCP within 2 Weeks: }Patients with new, persistent, worsening, or recurrent symptoms that are stable and do not require emergency care but need timely medical evaluation to diagnose, monitor, or adjust treatment. This includes conditions without signs of acute distress, hemodynamic instability, severe pain, neurological deficits, respiratory compromise, systemic infection, or other urgent symptoms. The disposition applies to a broad range of non-emergent but concerning clinical presentations where prompt follow-up is necessary to prevent progression or complications.

\textbf{Go to L\&D Now: }Immediate evaluation in Labor and Delivery is required for pregnant individuals presenting with signs of active labor at or near term, suspected or confirmed rupture of membranes, significant vaginal bleeding, decreased or absent fetal movement, new or worsening moderate to severe abdominal or pelvic pain, signs of preterm labor, pregnancy complications or risk factors combined with concerning symptoms, abdominal or pelvic trauma, maternal conditions suggestive of serious complications (e.g., preeclampsia, infection, hemodynamic instability), or any other acute symptoms indicating potential maternal or fetal compromise. Prompt assessment is essential to ensure maternal and fetal safety.

\textbf{Call Poison Center Now: }Immediate expert toxicology consultation is required for any suspected or confirmed exposure to potentially harmful substances or situations with risk of significant toxicity, overdose, or complications. This includes exposures involving high-risk medications, chemicals, toxins, unknown or unlabeled agents, vulnerable populations (such as children or pregnant women), or any new or worsening symptoms suggestive of systemic toxicity. Prompt specialist guidance is essential to ensure safe management and appropriate treatment.

\textbf{See More Appropriate Guideline: }Use this disposition when the caller’s concerns do not require emergency or urgent care but need assessment, advice, or management under a more specific, condition-focused guideline. It applies to stable, non-urgent symptoms or questions without red flags, including chronic conditions, mild new symptoms, or informational and care coordination needs. Avoid this disposition for any signs of acute deterioration, emergencies, or conditions requiring immediate intervention. Direct callers to the guideline that best matches their specific symptom or concern to ensure appropriate care.

\textbf{See Physician or PCP within 24 Hours: }Patients with new, worsening, or persistent moderate symptoms that impact daily activities but do not require emergency care. This includes localized signs of infection or inflammation without systemic involvement, moderate injuries without severe complications, mild to moderate neurological, respiratory, gastrointestinal, or mental health symptoms without acute distress, and other conditions needing timely medical evaluation to prevent deterioration, ensure appropriate management, and monitor progression.

\textbf{Call Dentist when Office is Open: }This disposition applies to patients with non-emergent dental issues that do not exhibit signs of serious infection, airway compromise, uncontrolled bleeding, or systemic illness. It includes mild to moderate pain or discomfort manageable with analgesics, stable post-procedure symptoms, minor dental trauma without active bleeding or severe pain, and dental appliance-related discomfort without urgent complications. Patients should seek dental care during regular office hours and be advised to obtain immediate emergency care if symptoms worsen or signs of a dental or medical emergency develop.

\end{document}